\documentclass[10pt,twocolumn,letterpaper]{article}

\usepackage{iccv}
\usepackage{times}
\usepackage{epsfig}
\usepackage{graphicx}
\usepackage{amsmath}
\usepackage{amssymb}
\usepackage{multirow}
\usepackage{pifont}

\usepackage[pagebackref=true,breaklinks=true,letterpaper=true,colorlinks,bookmarks=false]{hyperref}

\iccvfinalcopy 

\def\iccvPaperID{2278} 

\ificcvfinal\fi
\begin{document}

\title{LPRNet: License Plate Recognition via Deep Neural Networks}

\author{Sergey Zherzdev\\
ex-Intel\footnote{} \\
IOTG Computer Vision Group\\
{\tt\small sergeyzherzdev@gmail.com}
\and
Alexey Gruzdev\\
Intel \\
IOTG Computer Vision Group\\
{\tt\small alexey.gruzdev@intel.com}
}

\maketitle
\footnotetext{This work was done when Sergey was an Intel employee}

\begin{abstract}
   This paper proposes LPRNet - end-to-end method for Automatic License Plate Recognition without preliminary character segmentation. Our approach is inspired by recent breakthroughs in Deep Neural Networks, and works in real-time with recognition accuracy up to  {\bf 95\%} for Chinese license plates: {\bf 3 ms/plate} on nVIDIA\textsuperscript{\textregistered} GeForce\texttrademark GTX 1080 and {\bf 1.3 ms/plate} on Intel\textsuperscript{\textregistered} Core\texttrademark i7-6700K CPU.
   
   LPRNet consists of the  lightweight Convolutional Neural Network, so it can be trained in end-to-end way. To the best of our knowledge, LPRNet is the first real-time License Plate Recognition system that does not use RNNs. As a result, the LPRNet algorithm may be used to create embedded solutions for LPR that feature high level accuracy even on challenging Chinese license plates.
\end{abstract}

\section{Introduction}
\begin{figure*}
\begin{center}
\includegraphics[scale=0.5]{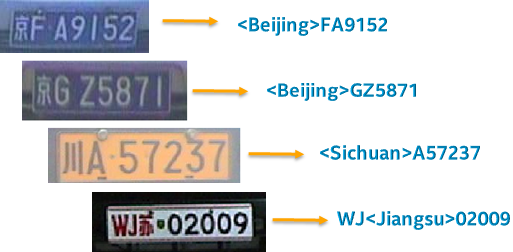}
\end{center}
   \caption{Example of LPRNet recognitions}
\label{fig:short}
\end{figure*}

Automatic License Plate Recognition is a challenging and important task which is used in traffic management, digital security surveillance, vehicle recognition, parking management of big cities.
This task is a complex problem due to many factors which include but are not limited to: blurry images, poor lighting conditions, variability of license plates numbers (including special characters \eg logograms for China, Japan), physical impact (deformations), weather conditions (see some examples in Fig. \ref{fig:short}).

The robust Automatic License Plate Recognition system needs to cope with a variety of environments while maintaining a high level of accuracy, in other words this system should work well in natural conditions.

This paper tackles the License Plate Recognition problem and introduces the LPRNet algorithm, which is designed to work without pre-segmentation and consequent recognition of characters. In the present paper, we do not consider License Plate Detection problem, however, for our particular case it can be done through LBP-cascade.

LPRNet is based on Deep Convolutional Neural Network. Recent studies proved effectiveness and superiority of Convolutional Neural Networks in many Computer Vision tasks such as image classification, object detection and semantic segmentation. However, running most of them on embedded devices still remains a challenging problem.

LPRNet is a very efficient neural network, which takes only \textbf{0.34 GFLops} to make a single forward pass. Also, our model is real-time on Intel® Core™ i7-6700K SkyLake CPU with high accuracy on challenging Chinese License plates and can be trained end-to-end. Moreover, LPRNet can be partially ported on FPGA, which can free up CPU power for other parts of the pipeline. Our main contributions can be summarized as follows:
\begin{itemize}
\item LPRNet is a real-time framework for high-quality license plate recognition supporting template and character independent variable-length license plates, performing LPR without
character pre-segmentation, trainable end-to-end from scratch for different national license plates.

\item LPRNet is the first real-time approach that does not use Recurrent Neural Networks and is lightweight enough to run on variety of platforms, including embedded devices.

\item Application of LPRNet to real traffic surveillance video shows that our approach is robust enough to handle difficult cases, such as perspective and camera-dependent distortions, hard lighting conditions, change of viewpoint, etc.
\end{itemize}

The rest of the paper is organized as follows. Section \ref{related} describes the related work. In sec. \ref{model} we review the details of the LPRNet model. Sec. \ref{results} reports the results on Chinese License Plates and includes an ablation study of our algorithm. We summarize and conclude our work in sec. \ref{conclusion}.

\section{Related Work}\label{related}

In the earlier works on general LP recognition, such as \cite{anagnostopoulos_license_2008} the pipeline consist of character segmentation and character classification stages:
\begin{itemize}

\item Character segmentation typically uses different hand-crafted algorithms, combining projections, connectivity and contour based image components. It takes a binary image or intermediate representation as input so character segmentation quality is highly affected by the input image noise, low resolution, blur or deformations.

\item Character classification typically utilizes one of the optical character recognition (OCR) methods -  adopted for LP character set.
\end{itemize}
Since classification follows the character segmentation, end-to-end recognition quality depends heavily on the applied segmentation method. In order to solve the problem of character segmentation there were proposed end-to-end Convolutional Neural Networks (CNNs) based solutions taking the whole LP image as input and producing the output character sequence.

The segmentation-free model in \cite{li_reading_2016} is based on variable length sequence decoding driven by connectionist temporal classification (CTC) loss \cite{graves_supervised_2012, graves_connectionist_2006}. It uses hand-crafted features LBP built on a binarized image as CNN input to produce character classes probabilities. Applied to all input image positions via the sliding window approach it makes the input sequence for the bi-directional Long-Short Term Memory (LSTM) \cite{hochreiter_long_1997} based decoder. Since the decoder output and target character sequence lengths are different, CTC loss is used for the pre-segmentation free end-to-end training.

The model in \cite{cheang_segmentation-free_2017} mostly follows the approach described in \cite{li_reading_2016} except that the sliding window method was replaced by CNN output spatial splitting to the RNN input sequence ("sliding window" over feature map instead of input).

In contrast \cite{jain_deep_2016} uses the CNN-based model for the whole LP image to produce the global LP embedding which is decoded to a 11-character-length sequence via 11 fully connected model heads. Each of the heads is trained to classify the i-th target string character (which is assumed to be padded to the predefined fixed length), so the whole recognition can be done in a single feed-forward pass. It also utilizes the Spatial Transformer Network (STN) \cite{jaderberg_spatial_2015} to reduce the effect of input image deformations.

The algorithm in \cite{end_to_end_lpdr} makes an attempt to solve both license plate detection and license plate recognition problems by single Deep Neural Network.

Recent work \cite{gans_data_lpr} tries to exploit synthetic data generation approach based on Generative Adversarial Networks \cite{gan_original} for data generation procedure to obtain large representative license plates dataset.

In our approach, we avoided using hand-crafted features over a binarized image - instead we used raw RGB pixels as CNN input. The LSTM-based sequence decoder working on outputs of a sliding window CNN was replaced with a fully convolutional model which output is interpreted as character probabilities sequence for CTC loss training and greedy or prefix search string inference. For better performance the pre-decoder intermediate feature map was augmented by the global context embedding as described in \cite{liu_parsenet:_2015}. Also the backbone CNN model was reduced significantly using the low computation cost basic building block inspired by SqueezeNet Fire Blocks \cite{iandola_squeezenet:_2016} and Inception Blocks of \cite{szegedy_inception-v4_2016, szegedy_going_2014, szegedy_rethinking_2015}. Batch Normalization \cite{ioffe_batch_2015} and Dropout \cite{srivastava_dropout:_2014} techniques were used for regularization.

LP image input size affects both the computational cost and the recognition quality \cite{agarwal_deciphering_2017}, as a result there is a trade-off between using high \cite{cheang_segmentation-free_2017} or moderate \cite{jain_deep_2016, li_reading_2016} resolution.

\section{LPRNet}\label{model}

\subsection{Design architecture}
In this section we describe our LPRNet network architecture design in detail.

In recent studies tend to use parts of the powerful classification networks such as VGG, ResNet or GoogLeNet as `backbone` for their tasks by applying transfer learning techniques.
However, this is not the best option for building fast and lightweight networks, so in our case we redesigned main `backbone` network applying recently discovered architecture tricks.

The basic building block of our CNN model backbone (Table \ref{basicblock})  was inspired by SqueezeNet Fire Blocks \cite{iandola_squeezenet:_2016} and Inception Blocks of \cite{szegedy_inception-v4_2016, szegedy_going_2014, szegedy_rethinking_2015}. We also followed the research best practices and used Batch Normalization \cite{ioffe_batch_2015} and ReLU activation after each convolution operation.

In a nutshell our design consists of:
\begin{itemize}
\item location network with Spatial Transformer Layer \cite{jaderberg_spatial_2015} (optional)
\item light-weight convolutional neural network (backbone)
\item per-position character classification head
\item character probabilities for further sequence decoding
\item post-filtering procedure
\end{itemize}

First, the input image is preprocessed by the Spatial Transformer Layer, as proposed in \cite{jaderberg_spatial_2015}. This step is optional but allows to explore how one can transform the input image to have better characteristics for recognition. The original LocNet (see the Table \ref{tab:locnet}) architecture was used to estimate optimal transformation parameters.

\begin{table}[h]
    \begin{tabular}{|l|c|c|}
        \hline
        Layer Type & \multicolumn{2}{c|}{Parameters} \\
        \hline\hline
        Input & \multicolumn{2}{c|}{94x24 pixels RGB image}  \\
        \hline
        AvgPooling &   \#32 3x3 stride 2 & \textemdash \\
        \hline
        Convolution &  \#32 5x5 stride 3 & \#32 5x5 stride 5 \\
        \hline
        Concatenation & \multicolumn{2}{c|}{channel-wise}  \\
        \hline
        Dropout & \multicolumn{2}{c|}{0.5 ratio}  \\
        \hline
        FC & \multicolumn{2}{c|}{\#32 with TanH  activation}  \\
        \hline
        FC & \multicolumn{2}{c|}{\#6 with scaled TanH activation}  \\
        \hline
    \end{tabular}
    \newline
    \caption{\textbf{LocNet architecture}}
    \label{tab:locnet}
\end{table}

\begin{table}[h]
\begin{tabular}{|l|c|}
\hline
Layer Type & Parameters/Dimensions \\
\hline\hline
Input & $C_{in} \times H \times W$ feature map \\
\hline
Convolution & \# $C_{out}/4$ 1x1 stride 1 \\
\hline
Convolution & \# $C_{out}/4$ 3x1 strideh=1, padh=1 \\
\hline
Convolution & \# $C_{out}/4$ 1x3 stridew=1, padw=1 \\
\hline
Convolution & \# $C_{out}$ 1x1 stride 1 \\
\hline
Output & $C_{out} \times H \times W$ feature map \\
\hline
\end{tabular}
\newline
\caption{\textbf{Small Basic Block}}
\label{basicblock}
\end{table}

The backbone network architecture is described in Table \ref{backbone}. The backbone takes a raw RGB image as input and calculates spatially distributed rich features. Wide convolution (with $1 \times 13$ kernel) utilizes the local character context instead of using LSTM-based RNN. The backbone subnetwork output can be interpreted as a sequence of character probabilities whose length corresponds to the input image pixel width. Since the decoder output and the target character sequence lengths are of different length, we apply the method of CTC loss \cite{hannun2017sequence} -  for segmentation-free end-to-end training. CTC loss is a well-known approach for situations where input and output sequences are misaligned and have variable lengths. Moreover, CTC provides an efficient way to go from probabilities at each time step to the probability of an output sequence. More detailed explanation about CTC loss can be found in .

\begin{table}[h]
\begin{tabular}{|l|c|}
\hline
Layer Type & Parameters  \\
\hline\hline
Input & 94x24 pixels RGB image  \\
\hline
Convolution & \#64 3x3 stride 1  \\
\hline
MaxPooling & \#64 3x3 stride 1 \\
\hline
Small basic block & \#128 3x3 stride 1 \\
\hline
MaxPooling & \#64 3x3 stride (2, 1) \\
\hline
Small basic block & \#256 3x3 stride 1 \\
\hline
Small basic block & \#256 3x3 stride 1 \\
\hline
MaxPooling & \#64 3x3 stride (2, 1) \\
\hline
Dropout & 0.5 ratio \\
\hline
Convolution & \#256 4x1 stride 1 \\
\hline
Dropout & 0.5 ratio \\
\hline
Convolution & \# class\_number 1x13 stride 1 \\
\hline
\end{tabular}
\newline
\caption{\textbf{Back-bone Network Architecture}}
\label{backbone}
\end{table}

To further improve performance, the pre-decoder intermediate feature map was augmented with the global context embedding as in \cite{liu_parsenet:_2015}. It is computed via a fully-connected layer over backbone output, tiled to the desired size and concatenated with backbone output. In order to adjust the depth of feature map to the character class number additional $1 \times 1$ convolution is applied.

For the decoding procedure at the inference stage we considered 2 options: greedy search and beam search. While greedy search takes the maximum of class probabilities in each position, beam search maximizes the total probability of the output sequence \cite{graves_supervised_2012, graves_connectionist_2006}.

For post-filtering we use a task-oriented language model implemented as a set of the target country LP templates. Note that post-filtering is applied together with Beam Search. The post-filtering procedure gets top-N most probable sequences found by beam search and returns the first one that matches the set of predefined templates which depends on country LP regulations.

\subsection{Training details}
All training experiments were done with the help of TensorFlow \cite{abadi_tensorflow:_2016}.

We train our model with '\textbf{Adam}' optimizer using batch size of \textbf{32}, initial learning rate \textbf{0.001} and gradient noise scale of \textbf{0.001}. We drop the learning rate once after every \textbf{100k} iterations by a factor of \textbf{10} and train our network for \textbf{250k} iterations in total.

In our experiments we use data augmentation by random affine transformations, \eg rotation, scaling and shift.

It is worth mentioning, that application of LocNet from the beginning of training leads to degradation of results, because LocNet cannot get reasonable gradients from a recognizer which is typically too weak for the first few iterations. So, in our experiments, we turn LocNet on only after \textbf{5k} iterations.

All other hyper-parameters were chosen by cross-validation over the target dataset.

\section{Results of the Experiments}\label{results}

The LPRNet baseline network, from which we started our experiments with different architectures, was inspired by \cite{li_reading_2016}. It's mainly based on Inception blocks followed by a bidirectional LSTM (biLSTM) decoder and trained with CTC loss. We first performed some experiments aimed at replacing biLSTM with biGRU cells, but we did not observe any clear benefits of using biGRU over biLSTM.

Then, we focused on eliminating of the complicated biLSTM decoder, because most modern embedded devices still do not have sufficient compute and memory to efficiently execute biLSTM. Importantly, our LSTM is applied to a spatial sequence rather than to a temporal one. Thus all LSTM inputs are known upfront both at the training stage as well as at the inference stage. Therefore we believe that RNN can be replaced by spatial convolutions without a significant drop in accuracy.
The RNN-less model with some backbone modifications is referenced as LPRNet basic and it was described in details in sec. \ref{model}.

To improve runtime performance we also modified LPRNet basic by using $2 \times 2$ strides for all pooling layers. This modification (the LPRNet reduced model) reduces the size of intermediate feature maps and total inference computational cost significantly (see GFLOPs column of the Table \ref{tab:results}).

\subsection{Chinese License Plates dataset}
We tested our approach on a private dataset containing various images of Chinese license plates collected from different security and surveillance cameras. This dataset was first run through the LPB-based detector to get bounding boxes for each license plate. Then, all license plates were manually labeled. In total, the dataset contains \textbf{11696} cropped license plate images, which are split as \textbf{9:1} into training and validation subsets respectively.

Automatically cropped license plate images were used for training to make the network more robust to detection artifacts because in some cases plates are cropped with some background around edges, while in other cases they are cropped too close to edges with no background at all or event with some parts of the license plate missing.

Table \ref{tab:results} shows recognition accuracies achieved by different models.

\begin{table}[h]
\begin{center}
\begin{tabular}{|l|c|c|}
\hline
Method & Recognition Accuracy, \% & GFLOPs\\
\hline\hline
LPRNet baseline & 94.1 & 0.71 \\
LPRNet basic & 95.0 & 0.34  \\
LPRNet reduced & 94.0 & 0.163 \\
\hline
\end{tabular}
\end{center}
\caption{\textbf{Results on Chinese License Plates.}}
\label{tab:results}
\end{table}

\subsection{Ablation study}
It is of vital importance to conduct the ablation study to identify correlation between various enhancements and respective accuracy/performance improvements. This helps other researchers adopt ideas from the paper and reuse most promising architecture approaches.
Table \ref{tab:ablation} shows a summary of architecture approaches and their impact on accuracy.

\begin{table}[h]
{\setlength{\tabcolsep}{0.2em}
\begin{tabular}{|l|cccccccc|}
\hline
Approach & \multicolumn{8}{c|}{LPRNet} \\
\hline\hline
Global context & & & & \ding{51} & \ding{51}  & \ding{51} & \ding{51} & \ding{51} \\
Data augm. & \ding{51} & \ding{51} & \ding{51} &  & \ding{51}  & \ding{51} & \ding{51} & \ding{51} \\
STN-alignment & & \ding{51} & \ding{51} & & & \ding{51} & \ding{51} & \ding{51} \\
Beam Search & & & \ding{51} &  &  &  & \ding{51} & \ding{51} \\
Post-filtering & & & \ding{51} &  &  &  &  & \ding{51} \\
\hline
Accuracy, \% & 53.4 & 58.6 & 59.0 & 62.95 & 91.6 & 94.4 & 94.4 & 95.0 \\
\hline
\end{tabular}}
\newline
\caption{\textbf{Effects of various tricks on LPRNet quality.}}
\label{tab:ablation}
\end{table}

As one can see, the largest accuracy gain ({\bf 36\%}) was achieved using the global context. The data augmentation techniques also help to improve accuracy significantly (by {\bf 28.6\%}). Without using data augmentation and the global context we could not train the model from scratch.

The STN-based alignment subnetwork provides noticeable improvement of 2.8-5.2\%. Beam Search with post-filtering further improves recognition accuracy by 0.4-0.6\%.

\subsection{Performance analysis}
The LPRNet reduced model was ported to various hardware platforms including CPU, GPU and FPGA. The results are presented in the Table \ref{tab:time}.

\begin{table}[h]
\begin{center}
\begin{tabular}{|l|c|}
\hline
Target platform & 1 LP processing time \\
\hline\hline
GPU + cuDNN & 3 ms  \\
\hline
CPU (using Caffe \cite{jia2014caffe}) & 11-15 ms  \\
\hline
CPU + FPGA (using DLA \cite{aydonat_opencltm_2017}) & 4 ms\footnote{} \\
\hline
CPU (using IE from Intel ® OpenVINO \cite{dlsdk}) & 1.3 ms \\
\hline
\end{tabular}
\end{center}
\caption{\textbf{Performance analysis.}}
\label{tab:time}\end{table}

\footnotetext{The LPRNet reduced model was used}

Here GPU is nVIDIA\textsuperscript{\textregistered} GeForce\texttrademark 1080, CPU is Intel\textsuperscript{\textregistered} Core\texttrademark i7-6700K SkyLake, FPGA is Intel\textsuperscript{\textregistered} Arria\texttrademark 10 and IE is for Inference Engine from   Intel\textsuperscript{\textregistered} OpenVINO.

\section{Conclusions and Future Work}\label{conclusion}
In this work, we have shown that for License Plate Recognition one can utilize pretty small convolutional neural networks. LPRNet model was introduced, which can be used for challenging data, achieving up \textbf{to 95\%} recognition accuracy. Architecture details, its motivation and the ablation study was conducted.

We showed that LPRNet can perform inference in real-time on a variety of hardware architectures including CPU, GPU and FPGA. We have no doubt that LPRNet could attain real-time performance even on more specialized embedded low-power devices.

The LPRNet can likely be compressed using modern pruning and quantization techniques, which would potentially help to reduce the computational complexity even further.


As a future direction of research, LPRNet work can be extended by merging CNN-based detection part into our algorithm, so that both detection and recognition tasks will be evaluated as a single network in order to outperform the LBP-based cascaded detector quality.

\section{Acknowledgements}\label{acknowledgements}
We would like to thank Intel\textsuperscript{\textregistered} IOTG Computer Vision (ICV) Optimization team for porting the model to the Intel ® Inference Engine of OpenVINO, as well as Intel\textsuperscript{\textregistered} IOTG Computer Vision (ICV) OVX FPGA team for porting the model to the DLA.
We also would like to thank Intel\textsuperscript{\textregistered} PSG DLA and Intel\textsuperscript{\textregistered} Computer Vision SDK teams for their tools and support.


{\small
\bibliographystyle{IEEEtran}
\bibliography{lprnet}

\begin{thebibliography}{10}
\providecommand{\url}[1]{#1}
\csname url@samestyle\endcsname
\providecommand{\newblock}{\relax}
\providecommand{\bibinfo}[2]{#2}
\providecommand{\BIBentrySTDinterwordspacing}{\spaceskip=0pt\relax}
\providecommand{\BIBentryALTinterwordstretchfactor}{4}
\providecommand{\BIBentryALTinterwordspacing}{\spaceskip=\fontdimen2\font plus
\BIBentryALTinterwordstretchfactor\fontdimen3\font minus
  \fontdimen4\font\relax}
\providecommand{\BIBforeignlanguage}[2]{{%
\expandafter\ifx\csname l@#1\endcsname\relax
\typeout{** WARNING: IEEEtran.bst: No hyphenation pattern has been}%
\typeout{** loaded for the language `#1'. Using the pattern for}%
\typeout{** the default language instead.}%
\else
\language=\csname l@#1\endcsname
\fi
#2}}
\providecommand{\BIBdecl}{\relax}
\BIBdecl

\bibitem{anagnostopoulos_license_2008}
C.~N.~E. Anagnostopoulos, I.~E. Anagnostopoulos, I.~D. Psoroulas, V.~Loumos,
  and E.~Kayafas, ``License {Plate} {Recognition} {From} {Still} {Images} and
  {Video} {Sequences}: {A} {Survey},'' \emph{IEEE Transactions on Intelligent
  Transportation Systems}, vol.~9, no.~3, pp. 377--391, Sep. 2008.

\bibitem{li_reading_2016}
H.~Li and C.~Shen, ``Reading {Car} {License} {Plates} {Using} {Deep}
  {Convolutional} {Neural} {Networks} and {LSTMs},'' \emph{arXiv:1601.05610
  [cs]}, Jan. 2016, arXiv: 1601.05610.

\bibitem{graves_supervised_2012}
A.~Graves, \emph{\BIBforeignlanguage{English}{Supervised {Sequence} {Labelling}
  with {Recurrent} {Neural} {Networks}}}, 2012th~ed.\hskip 1em plus 0.5em minus
  0.4em\relax Heidelberg ; New York: Springer, Feb. 2012.

\bibitem{graves_connectionist_2006}
A.~Graves, S.~Fernández, F.~Gomez, and J.~Schmidhuber, ``Connectionist
  temporal classification: labelling unsegmented sequence data with recurrent
  neural networks,'' in \emph{Proceedings of the 23rd international conference
  on {Machine} learning}.\hskip 1em plus 0.5em minus 0.4em\relax ACM, 2006, pp.
  369--376.

\bibitem{hochreiter_long_1997}
S.~Hochreiter and J.~Schmidhuber, ``\BIBforeignlanguage{en}{Long {Short}-{Term}
  {Memory}},'' \emph{\BIBforeignlanguage{en}{Neural Computation}}, vol.~9,
  no.~8, pp. 1735--1780, Nov. 1997.

\bibitem{cheang_segmentation-free_2017}
T.~K. Cheang, Y.~S. Chong, and Y.~H. Tay, ``Segmentation-free {Vehicle}
  {License} {Plate} {Recognition} using {ConvNet}-{RNN},''
  \emph{arXiv:1701.06439 [cs]}, Jan. 2017, arXiv: 1701.06439.

\bibitem{jain_deep_2016}
V.~Jain, Z.~Sasindran, A.~Rajagopal, S.~Biswas, H.~S. Bharadwaj, and K.~R.
  Ramakrishnan, ``Deep {Automatic} {License} {Plate} {Recognition} {System},''
  in \emph{Proceedings of the {Tenth} {Indian} {Conference} on {Computer}
  {Vision}, {Graphics} and {Image} {Processing}}, ser. {ICVGIP} '16.\hskip 1em
  plus 0.5em minus 0.4em\relax New York, NY, USA: ACM, 2016, pp. 6:1--6:8.

\bibitem{jaderberg_spatial_2015}
M.~Jaderberg, K.~Simonyan, A.~Zisserman, and K.~Kavukcuoglu, ``Spatial
  {Transformer} {Networks},'' \emph{arXiv:1506.02025 [cs]}, Jun. 2015, arXiv:
  1506.02025.

\bibitem{end_to_end_lpdr}
H.~{Li}, P.~{Wang}, and C.~{Shen}, ``{Towards End-to-End Car License Plates
  Detection and Recognition with Deep Neural Networks},'' \emph{ArXiv
  e-prints}, Sep. 2017.

\bibitem{gans_data_lpr}
X.~{Wang}, Z.~{Man}, M.~{You}, and C.~{Shen}, ``{Adversarial Generation of
  Training Examples: Applications to Moving Vehicle License Plate
  Recognition},'' \emph{ArXiv e-prints}, Jul. 2017.

\bibitem{gan_original}
I.~J. {Goodfellow}, J.~{Pouget-Abadie}, M.~{Mirza}, B.~{Xu}, D.~{Warde-Farley},
  S.~{Ozair}, A.~{Courville}, and Y.~{Bengio}, ``{Generative Adversarial
  Networks},'' \emph{ArXiv e-prints}, Jun. 2014.

\bibitem{liu_parsenet:_2015}
W.~Liu, A.~Rabinovich, and A.~C. Berg, ``{ParseNet}: {Looking} {Wider} to {See}
  {Better},'' \emph{arXiv:1506.04579 [cs]}, Jun. 2015, arXiv: 1506.04579.

\bibitem{iandola_squeezenet:_2016}
F.~N. Iandola, S.~Han, M.~W. Moskewicz, K.~Ashraf, W.~J. Dally, and K.~Keutzer,
  ``{SqueezeNet}: {AlexNet}-level accuracy with 50x fewer parameters and
  {\textless}0.5mb model size,'' \emph{arXiv:1602.07360 [cs]}, Feb. 2016,
  arXiv: 1602.07360.

\bibitem{szegedy_inception-v4_2016}
C.~Szegedy, S.~Ioffe, V.~Vanhoucke, and A.~Alemi, ``Inception-v4,
  {Inception}-{ResNet} and the {Impact} of {Residual} {Connections} on
  {Learning},'' \emph{arXiv:1602.07261 [cs]}, Feb. 2016, arXiv: 1602.07261.

\bibitem{szegedy_going_2014}
C.~Szegedy, W.~Liu, Y.~Jia, P.~Sermanet, S.~Reed, D.~Anguelov, D.~Erhan,
  V.~Vanhoucke, and A.~Rabinovich, ``Going {Deeper} with {Convolutions},''
  \emph{arXiv:1409.4842 [cs]}, Sep. 2014, arXiv: 1409.4842.

\bibitem{szegedy_rethinking_2015}
C.~Szegedy, V.~Vanhoucke, S.~Ioffe, J.~Shlens, and Z.~Wojna, ``Rethinking the
  {Inception} {Architecture} for {Computer} {Vision},'' \emph{arXiv:1512.00567
  [cs]}, Dec. 2015, arXiv: 1512.00567.

\bibitem{ioffe_batch_2015}
S.~Ioffe and C.~Szegedy, ``Batch {Normalization}: {Accelerating} {Deep}
  {Network} {Training} by {Reducing} {Internal} {Covariate} {Shift},''
  \emph{arXiv:1502.03167 [cs]}, Feb. 2015, arXiv: 1502.03167.

\bibitem{srivastava_dropout:_2014}
N.~Srivastava, G.~Hinton, A.~Krizhevsky, I.~Sutskever, and R.~Salakhutdinov,
  ``Dropout: {A} {Simple} {Way} to {Prevent} {Neural} {Networks} from
  {Overfitting},'' \emph{J. Mach. Learn. Res.}, vol.~15, no.~1, pp. 1929--1958,
  Jan. 2014.

\bibitem{agarwal_deciphering_2017}
S.~Agarwal, D.~Tran, L.~Torresani, and H.~Farid, ``Deciphering {Severely}
  {Degraded} {License} {Plates},'' San Francisco, CA, 2017.

\bibitem{hannun2017sequence}
A.~Hannun, ``Sequence modeling with ctc,'' \emph{Distill}, 2017,
  https://distill.pub/2017/ctc.

\bibitem{abadi_tensorflow:_2016}
M.~Abadi, A.~Agarwal, P.~Barham, E.~Brevdo, Z.~Chen, C.~Citro, G.~S. Corrado,
  A.~Davis, J.~Dean, M.~Devin, S.~Ghemawat, I.~Goodfellow, A.~Harp, G.~Irving,
  M.~Isard, Y.~Jia, R.~Jozefowicz, L.~Kaiser, M.~Kudlur, J.~Levenberg, D.~Mane,
  R.~Monga, S.~Moore, D.~Murray, C.~Olah, M.~Schuster, J.~Shlens, B.~Steiner,
  I.~Sutskever, K.~Talwar, P.~Tucker, V.~Vanhoucke, V.~Vasudevan, F.~Viegas,
  O.~Vinyals, P.~Warden, M.~Wattenberg, M.~Wicke, Y.~Yu, and X.~Zheng,
  ``{TensorFlow}: {Large}-{Scale} {Machine} {Learning} on {Heterogeneous}
  {Distributed} {Systems},'' \emph{arXiv:1603.04467 [cs]}, Mar. 2016, arXiv:
  1603.04467.

\bibitem{jia2014caffe}
Y.~Jia, E.~Shelhamer, J.~Donahue, S.~Karayev, J.~Long, R.~Girshick,
  S.~Guadarrama, and T.~Darrell, ``Caffe: Convolutional architecture for fast
  feature embedding,'' \emph{arXiv preprint arXiv:1408.5093}, 2014.

\bibitem{aydonat_opencltm_2017}
U.~Aydonat, S.~O'Connell, D.~Capalija, A.~C. Ling, and G.~R. Chiu, ``An
  {OpenCL}({TM}) {Deep} {Learning} {Accelerator} on {Arria} 10,''
  \emph{arXiv:1701.03534 [cs]}, Jan. 2017, arXiv: 1701.03534.

\bibitem{dlsdk}
\BIBentryALTinterwordspacing
``{Intel OpenVINO Toolkit} {\textbar} {Intel}® {Software}.'' [Online].
  Available:
  \url{https://software.intel.com/en-us/articles/OpenVINO-InferEngine}
\BIBentrySTDinterwordspacing

\end{thebibliography}
}

\end{document}